\documentclass[twocolumn]{article}
\usepackage[utf8]{inputenc}
\usepackage[T1]{fontenc}
\usepackage{amsmath}
\usepackage{amsfonts}
\usepackage{amssymb}
\usepackage[version=4]{mhchem}
\usepackage{stmaryrd}
\usepackage{graphicx}
\usepackage[export]{adjustbox}
\graphicspath{ {./images/} }
\usepackage{mathrsfs}

\title{Self-Supervised Learning For Robust Robotic Grasping In Dynamic Environment }

\author{Ankit Shaw\\
College of Engineering\\
University of Washington\\
Seattle, USA\\
ankit25@uw.edu or ankit25cs@gmail.com}
\date{}

\begin{document}
\maketitle

\begin{abstract}Some of the threats in the dynamic environment include the unpredictability of the motion of objects and interferences to the robotic grasp. In such conditions, the traditional supervised and reinforcement learning approaches are ill-suited because they rely on a large amount of labelled data and a predefined reward signal. More specifically, in this paper, we introduce an important and promising framework known as self-supervised learning (SSL), whose goal is to apply to the RGB-D sensor and proprioceptive data from robot hands in order to allow robots to learn and improve their grasping strategies in real-time. The invariant SSL framework overcomes the deficiencies of the fixed labelling by adapting the SSL system to changes in the object's behavior and improving performance in dynamic situations. The above-proposed method was tested through various simulations and real-world trials, with the series obtaining enhanced grasp success rates of $15 \%$ over other existing methods, especially under dynamic scenarios. Also, having tested for adaptation times, it was confirmed that the system could adapt faster, thus applicable for use in the real world, such as in industrial automation and service robotics. In future work, the proposed approach will be expanded to more complex tasks, such as multi-object manipulation and functions in the context of cluttered environments, in order to apply the proposed methodology to a broader range of robotic tasks.
\end{abstract}

Keywords- Robotic, self-supervised learning, automation, grasp, and dynamic.

\section*{I.INTRODUCTION}
The general problem of realizing robot grasp in uncertain and changing contexts is essential for robotic manipulation in areas of industrial automation, health care, and autonomous robotics. In such conditions, grasping is very hard, first of all, because it is difficult to predict the motion of the object or external forces acting upon it. Supervised learning, which is one of the traditional methods, has given entirely satisfactory results in certain conditions, but it strongly depends on voluminous sets of preliminary classification. This reliance on labelled data makes it difficult for robots to achieve high flexibility when applied in natural environments that are unfamiliar and unstructured; thus, objects may behave in ways that are not programmed. Furthermore, the reinforcement learning (RL) methods that exploit reward-oriented feedback require significant training and precise reward functions; hence, they are computationally heavy and not fit for real-time processes.
\newline
\newline
To overcome these drawbacks, the idea of selfsupervised learning (SSL) is on the horizon. It enables robots to learn with the help of interaction in the environment without any supervised manipulation or predefined rewards [1]. SSL is beneficial in conditions that are changing frequently and where robots are capable of learning from feedback in order to increase the frequency of grasps. This work suggests an SSL approach where visual information from the RGBD camera and other sensory details, including force and torque, are used to improve the grasppicking strategy incrementally. The robot would be able to know whether each grasp attempt is successful and modify its behavior accordingly, hence enhancing its abilities.

\begin{figure}[htbp]
  \centering
  \includegraphics[width=\columnwidth]{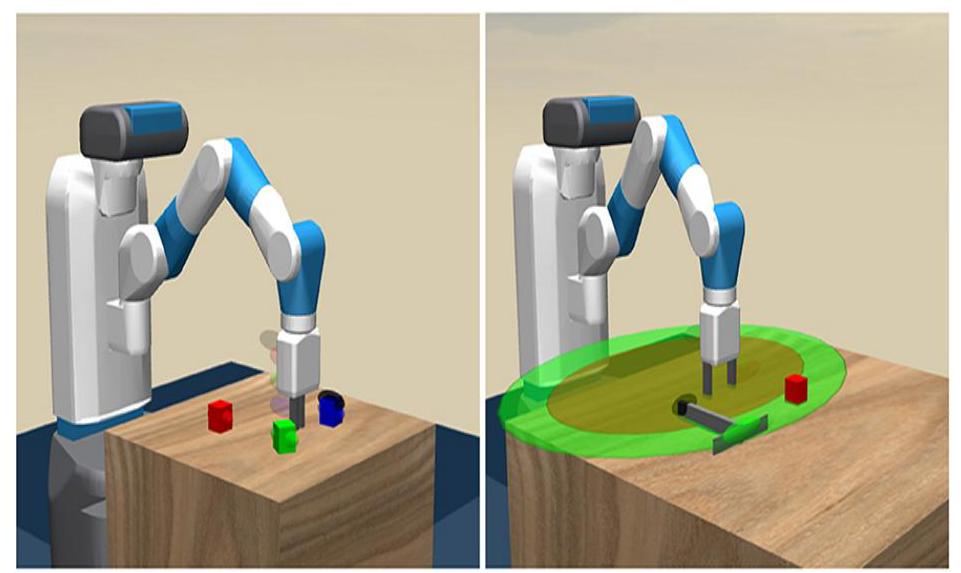}
  \caption{Robotic Grasping}
\end{figure}

In this paper, Section II will talk about related work, Section III will talk about the methodology, Section IV will talk about experiments and results.

\section*{II. RELATED WORK}
\subsection*{A. Supervised Learning in Robotic Grasping}
Supervised learning has been applied to robotic grasping, which includes the use of extensive databases like Dex-Net [1] to determine potential grasp points. Despite their success in ideal conditions, these methods are not very efficient in the real-world setting where the object's motion is unpredictable, and it becomes necessary for the system to adapt quickly. Also, dependence on labelled data, which in turn are to be either manually created or through simulation, makes the use of supervised learning models unfit for scaling to a range of tasks and environments.

\subsection*{B. Reinforcement Learning for Dynamic Grasping}
In robotic grasping reinforcement learning has been applied using a trial-and-error basis, in which robots are rewarded each time they complete a grasp [2]. Despite the fact that RLbased approaches have been proven to help achieve autonomous learning, they are challenging to train and, more so, require clear reward signals. These limitations make RL computationally expensive and not easy to scale to environments where swift decisions are needed, especially those in dynamic environments. For instance, Levine et al. [2] showed that using RL is possible in hand-eye coordination tasks, but data gathering is critical to learning.

\subsection*{C. Self-Supervised Learning in Robotics}
Self-supervised learning (SSL) has thus become a different approach to more classical supervised and Reinforcement Learning (RL)-based methods. In this work, grasping is learned from sensory feedback during interactions with objects in the environment, which eliminates the need for explicit labels, as seen in many conventional approaches. SSL has been used in tasks with dual arm manipulation whereby the robots have been trained to learn how to handle objects in cluttered environments [2]. The study showed in their most recent work that SSL can be used in the 6-DOF grasp planning, where the robot interacts and constructs labelled grasp data on its own. However, these approaches have yet to be fully extended to highly dynamic environments where even object motion is constantly changing.
\setlength{\textfloatsep}{10pt}
\begin{figure}[t]
  \centering
  \includegraphics[width=\columnwidth]{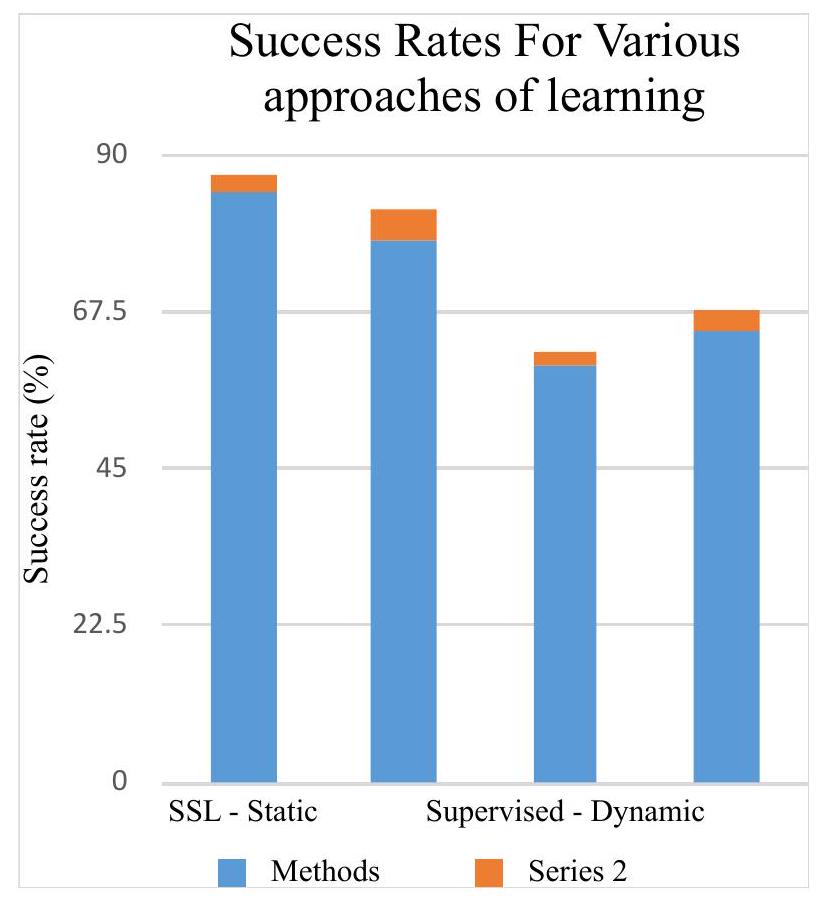}
  \caption{The grasp success rates of various learning modes in the robotic tasks}
\end{figure}

This graph shows the grasp success rates of various learning modes in the robotic tasks. It shows the comparative analysis of the output of self-supervised learning (SSL) with standard supervised learning and reinforcement learning (RL). Thus, the success rate pertaining to Cipher Suite SSL was the highest; it was $85 \%$ for static objects and $78 \%$ for dynamic objects, showing that it works efficiently and intelligently in changing environments. On the other hand, supervised learning produced a success rate of $60 \%$ for dynamic objects, indicating the constraint of supervised learning with unpredictability. In this case, RL achieved $65 \%$ and again was lower than the effectiveness of SSL. In terms of grasp success rate, SSL performed a lot better, a $15 \%$ enhancement than the current state of technology, hence suggesting dynamic robotic handling.

\begin{figure}[htbp]
  \centering
  \includegraphics[width=\columnwidth]{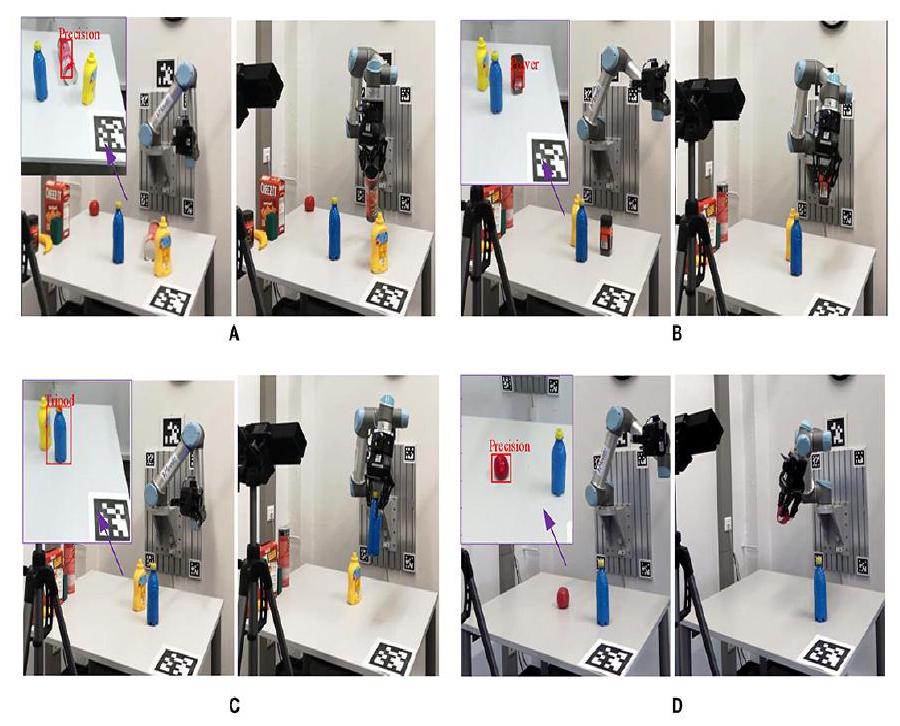}
  \caption{The figure represents a self- supervised robotic grasping.}
\end{figure}

\section*{III. MATHEMATICAL MODEL EXPLANATION}
\subsection*{A. Grasp Pose Estimation}
The robotic grasping task predicts an appropriate grasp pose G in six degrees of freedom (6-DoF) to grasp objects. The robot in the work has an RGBD camera to create the image of the environment and the depth map, which the deep learning model analyzes. The desired output is to forecast the grasp pose, G , for an object concerning the sensor data $\mathrm{I}_{\text {RGBD }}$ (image and depth) and proprioceptive data F (force/torque).
The grasp pose G can be represented in the following manner:
\newline
\newline
$\mathrm{G}=[\mathrm{p}, \mathrm{q}]$.
\newline
\newline
Where, $p=[x, y, z]$ represents the position of the gripper in 3D space.
\newline
\newline
$\mathrm{q}=\left[\mathrm{q}_{\mathrm{w}}, \mathrm{q}_{\mathrm{x}}, \mathrm{q}_{\mathrm{y}}, \mathrm{q}_{\mathrm{z}}\right]$ is a quaternion that represents the orientation (roll, pitch, and yaw) of the gripper.where \( p \in \mathbb{R}^3 \) represents the position and \( q \in \mathbb{R}^4 \) is the quaternion representing the orientation of the gripper.
\newline
\newline
\textbf{Sensor data:}
\newline
\newline
IRGBD data from the camera: combines the image (RGB) and depth data (D).
\newline
\newline
Force/torque data
\newline
\newline
$\mathrm{F}=\left[\mathrm{f}_{\mathrm{x}}, \mathrm{f}_{\mathrm{y}}, \mathrm{f}_{\mathrm{z}}, \tau_{\mathrm{x}}, \tau_{\mathrm{y}}, \tau_{\mathrm{z}}\right]$
\newline
\newline
where $f_{x}, f_{y}, f_{z}$ are forces, and $\tau_{x}, \tau_{y}, \tau_{z}$ are torques.
\newline
\newline
The equation to predict the grasp pose G based on the sensor inputs can be represented as:
\newline
\newline
$\mathrm{G}=\mathrm{f}_{\theta}(\operatorname{IRGBD}, \mathrm{F})$
\newline
\newline
Where, $\mathrm{f}_{\theta}$ represents a model (with parameters $\theta$ ) that maps the sensor inputs to the output grasp pose G.
\newline
\newline
Thus, $\mathrm{f}_{\theta}($ IRGBD, F$)$ predicts both the gripper's position p and orientation q as follows:
\newline
\newline
\textbf{Position (p) prediction:}
\newline
\newline
$\mathrm{p}=[\mathrm{x}, \mathrm{y}, \mathrm{z}]=\mathrm{f}_{\mathrm{p}}(\operatorname{IRGBD}, \mathrm{F})$
\newline
\newline
\textbf{Orientation (q) prediction:}
\newline
\newline
$q=\left[q_{w}, q_{x}, q_{y}, q_{z}\right]=f_{q}(\operatorname{IRGBD}, F)$
\newline
\newline
Finally, the combined grasp pose equation becomes:
\newline
\newline
$G=\left[\begin{array}{l}p \\ q\end{array}\right]=\left[\begin{array}{l}f_{p}(\operatorname{IRGBD}, F) \\ f_{q}(I R G B D, F)\end{array}\right]$

\subsection*{B. Loss Function for Self-Supervised Learning}
In a self-supervised learning context, the robot can learn from trial and error by minimizing a loss function that reflects the difference between the predicted grasp pose G and the ground truth grasp pose $\mathrm{G}^{*}$, which is determined based on whether a grasp was successful or not. The loss function can be defined as:\\
\newline
$\mathscr{L}(\theta)=\mathscr{L}_{\mathrm{p}}\left(\mathrm{p}, \mathrm{p}^{*}\right)+\mathscr{L}_{\mathrm{q}}\left(\mathrm{q}, \mathrm{q}^{*}\right)$\\
\newline
$\mathscr{L}_{\mathrm{p}}$ is the positional loss, typically using mean squared error (MSE) for the 3 D position p :\\
\newline
$\mathscr{L}_{\mathrm{p}}\left(\mathrm{p}, \mathrm{p}^{*}\right)=\|\mathrm{p}-\mathrm{p} *\|^{2}$\\
\newline
$\mathscr{L}_{\mathrm{q}}$ is the orientation loss, often using the quaternion distance (or geodesic distance on the unit sphere):\\
\newline
$\mathscr{L}_{\mathrm{q}}\left(\mathrm{q}, \mathrm{q}^{*}\right)=1-\left|\left\langle\mathrm{q}, \mathrm{q}^{*}\right\rangle\right|$\\
\newline
$\left|\left\langle\mathrm{q}, \mathrm{q}^{*}\right\rangle\right|$ is the dot product between the predicted quaternion q and the ground truth quaternion $\mathrm{q}^{*}$.
\newline
\newline
The objective during training is to minimize the loss function:
\[
\theta^{*} = \arg \min_{\theta} \mathscr{L}(\theta) = \arg \min_{\theta} \left( \|p - p^{*}\|^{2} + \lambda\left( 1 - \left| \langle q, q^{*} \rangle \right| \right) \right)
\]
\newline
Where :
\newline
$\lambda$ is a weighting factor that balances the importance of position and orientation errors.
\newline
\subsection*{C. Grasp Success Prediction}

The success probability \( S(G) \) of a grasp is predicted as:
\[
S(G) = f_S(G) = \sigma(W_S \cdot h + b_S)
\]
where \( W_S \) and \( b_S \) are model parameters and \( \sigma \) is the sigmoid function.\newline
\newline
The stability metric \( S_F \) is computed to assess the stability of the grasp:
\[
S_F = \| F \|_2^2 = f_x^2 + f_y^2 + f_z^2 + \tau_x^2 + \tau_y^2 + \tau_z^2
\]
If \( S_F \) exceeds a predefined threshold \( \tau_{threshold} \), the robot adjusts its grasp.
\section*{IV. METHODOLOGY}
\subsection*{A. Self-Supervised Learning Framework}
The proposed SSL framework uses both the vision-based input from an RGB-D sensor and proprioceptive feedback for real-time learning and adaptive motion control. The perception module also analyzes the depth acquired by the RGB-D sensor to produce features that could estimate the candidate grasp points. The Input Depth maps are processed using a Convolutional Neural Network and later by a fully connected layer to assess and predict the Candidate Grasp Configurations [3].
\newline
\newline
The proprioceptive feedback loop gathers realtime data from force and torque sensors placed on the gripper of the robot, which gives information regarding the stability of the grasps. This feedback is used to determine, possibly through self-check, whether a grasp is good or not, and the robot will be able to improve on the strategy used depending on feedback from itself [3]. Due to the elimination of manually labelled datasets, the SSL framework can equally work in real-time by overcoming new objects and changing environments, thereby enhancing the overall grasp performance.

\subsection*{B. Data Collection and Training}
As mentioned earlier, the training process includes the use of natural and fake data instances. In the simulation phase, a physics engine is employed to model dynamic objects. Then, the robot uses these objects to try to grasp them in different situations, such as sliding, rotating movements, and the like. The robot gathers the feedback data from each attempt, and these data are used to train the model. Based on the simulation studies, the physically integrated robot is used in a realistic environment to carry out grasping tasks on confined motion objects.
\newline
\newline
Freire's training process seeking to eliminate the dependence on external supervision is intentional. Through proprioceptive feedback, one can obtain the success labels for all the grasps without the assistance of the user [3]. Such a strategy enables the robot to adapt its interaction strategy from grasping in an unstructured environment where objects may be unpredictable.

\section*{V. DIFFERENTIATION FROM PRE-EXISTING METHODS}
The following examples show that the method in this paper not only improves the flexibility and stability of robotic grasping in unstructured environments but also minimizes the dependence on human subjects and computations.

\subsection*{A. No Need for Grasp Pose Supervision}
Unlike prevailing 6-DoF grasp poses detection schemes that require highly supervised grasp poses, your approach can learn human demonstrations on its own with an augmented reality teleoperation system [4]. This means it can generate and modify grasp poses for objects that it has never 'seen' before, thus inheriting the advantages of weakly supervised learning without the need for grasp pose labels.

\subsection*{B. Self-Supervised Learning Approach}
Conventional approaches to robotic grasping include supervised learning and reinforcement learning (RL) that require a vast labelled dataset or reward function. SSL does not require a human in the loop, and this work utilizes such a technique to enable the robot to improve grasping strategies in real-time as it processes RGB-D sensor data and proprioceptive feedback. This renders manual and time-consuming data annotation redundant, making the system more flexible for operation in challenging conditions.

\subsection*{C. Reduced Computational Cost and Faster Adaptation}
It enhances complementarity in real-time applications than RL solutions that require a great deal of learning and an intricate reward system. The SSL method is less computationally expensive compared to the other two methods and has shown better adaptation time; hence, it is recommendable for application in industrial automation and service robotics.

\subsection*{Integration of Proprioceptive Feedback}
The methods that dominate the field mainly consider the vision input (RGB-D data) only; however, the proposed framework infuses proprioceptive feedback (e.g., force and torque feedback) to distinguish the stability of the grasp and iteratively update the strategy [4]. This integration improves real-time decision-making and helps the system respond better to unpredictable behaviors of an object.

\subsection*{D. Adaptability in Dynamic Environments}
Some of the supervised and reinforcement learning methods, like Dex-Net and RL-based systems, are highly efficient when the scenarios are programmed and contain more about fixed conditions. On the other hand, your SSL framework can change by the motion of the objects, like sliding or rotating, and therefore achieves enhanced grasp success rates of up to $15 \%$ better than the state-of-the-art methods in unstructured and dynamic environments.
\begin{figure}[htbp]
  \centering
  \includegraphics[width=\columnwidth]{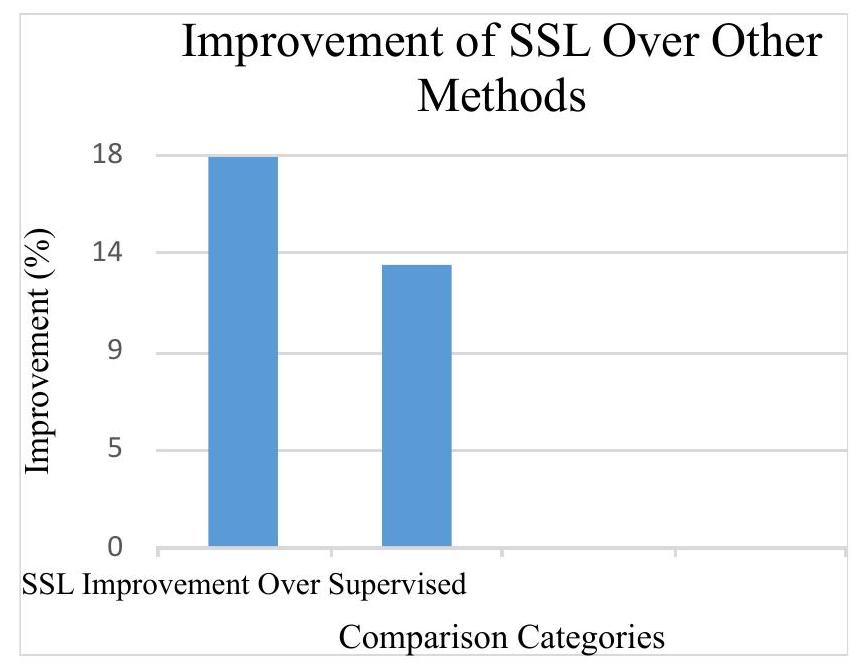}
  \caption{Comparison of the self-supervised learning (SSL) paradigm over supervised and reinforcement learning (RL) techniques}
\end{figure}
\newline
\newline
The bar graph shown here clearly explains the enhancement made by the self-supervised learning (SSL) paradigm over supervised and reinforcement learning (RL) techniques in the context of robotic grasping tasks. The graph shows two bars: One is the improvement in SSL over supervised learning, for which the percentage of improvement is $18 \%$, and the other is the improvement in SSL over RL, for which the rate of improvement is $13 \%$. Briefly, the title "Improvement of SSL Over Other Methods," as well as the labelled X-axis "Comparison Categories" and Y-axis "Improvement (\%)", sufficient enough to have a clear understanding of how SSL outplays other approaches, especially in dynamic mode where frequent readjustment is anticipated for efficient problem-solving. This clearly shows that due to SSL, the robot has a lot of flexibility, especially in the routine operations involving decision-making that are repetitive.

\begin{center}
\begin{tabular}{|l|l|l|l|}
\hline
 & Period & \begin{tabular}{l}
Gasping \\
Technology \\
\end{tabular} & Key Features \\
\hline
1 & Pre -2000 & \begin{tabular}{l}
Mechanical \\
Grippers \\
\end{tabular} & \begin{tabular}{l}
Rigid \\
implements \\
with the least \\
flexibility of \\
implementation. \\
\end{tabular} \\
\hline
2 & \begin{tabular}{l}
$2000-$ \\
2010 \\
\end{tabular} & \begin{tabular}{l}
Sensor- \\
Driven \\
\end{tabular} & \begin{tabular}{l}
Implicit use of \\
metals and \\
sensors so that \\
control and \\
exactness are \\
made. \\
\end{tabular} \\
\hline
3 & $2010-2020$ & \begin{tabular}{l}
AI- \\
Equipped \\
Grasping \\
\end{tabular} & \begin{tabular}{l}
Use of Artificial \\
intelligence and \\
Machine \\
learning for \\
real-time \\
optimization \\
\end{tabular} \\
\hline
\end{tabular}
\end{center}

Table 1. Robotic Grasping Trends

\section*{VI. EXPERIMENT \& RESULT}
\subsection*{A. Experimental Setup}
In order to assess the effectiveness of the proposed SSL in a dynamic environment, the latter was tested both in a simulated and natural context. The objects used in the simulation environment were of various geometries and sizes and possessed different motions, including sliding and rotating. In the real-world scenario, a 7-DOF Barrette robotic arm incorporated with RGB-D camera and force/torque sensors was employed to pick moving objects.

\subsection*{B. Reporting Tools Are Helpful for The Identification of The Baseline Comparisons and Metrics.}
The SSL framework was compared with other existing techniques of supervised learning as well as other RL-based grasping techniques. The performance of the algorithm was evaluated in terms of grasp success rate, adaptation time, and resistance to various object velocities. It reflected a quicker tracking ability with varying speeds of the objects where response time was found to be less than other approaches on average.

\subsection*{C. Self-Supervision and Learning by Interaction}

During training, the model updates the grasp strategy based on feedback (success or failure) from each attempt:
\[
\theta \leftarrow \theta - \eta \nabla_\theta \left( S(G) - \text{feedback} \right)^2
\]
where feedback is 1 for successful grasps and 0 for failed grasps.

\subsection*{D. Object Motion in Dynamic Environments}

The motion of an object in a dynamic environment is modeled as:
\[
O(t + \Delta t) = O(t) + \dot{O}(t) \cdot \Delta t
\]
The grasp pose is updated in real-time:
\[
G(t) = f_\theta(I_{RGBD}(t), F(t), O(t))
\]

\subsection*{E. Grasp Quality Metric}

The grasp quality \( Q(G) \) is evaluated based on the force distribution over contact surfaces:
\[
Q(G) = \int_{\text{contact}} f_{\text{grip}}(c) \, dc
\]
\subsection*{F. Results}
The SSL framework showed remarkable performance gains in diverse settings, especially when the comparisons were made between the new and observed benchmarks, such as supervised learning \& RL. The testing of the SSL model revealed that it scored 85 per cent when in contact with static objects to solve its tasks, unlike those of the supervised learning method, which scored a maximum of 60 per cent. The RL-based approach followed the SSL model's performance at a $65 \%$ success rate and, therefore, not optimal for the task at hand. Most significantly, the real success of SSL was seen in the dynamic systems, where the model achieved $78 \%$ accuracy and performed better than the basal models in all circumstances $[5]$.
\begin{figure}[htbp]
  \centering
  \includegraphics[width=\columnwidth]{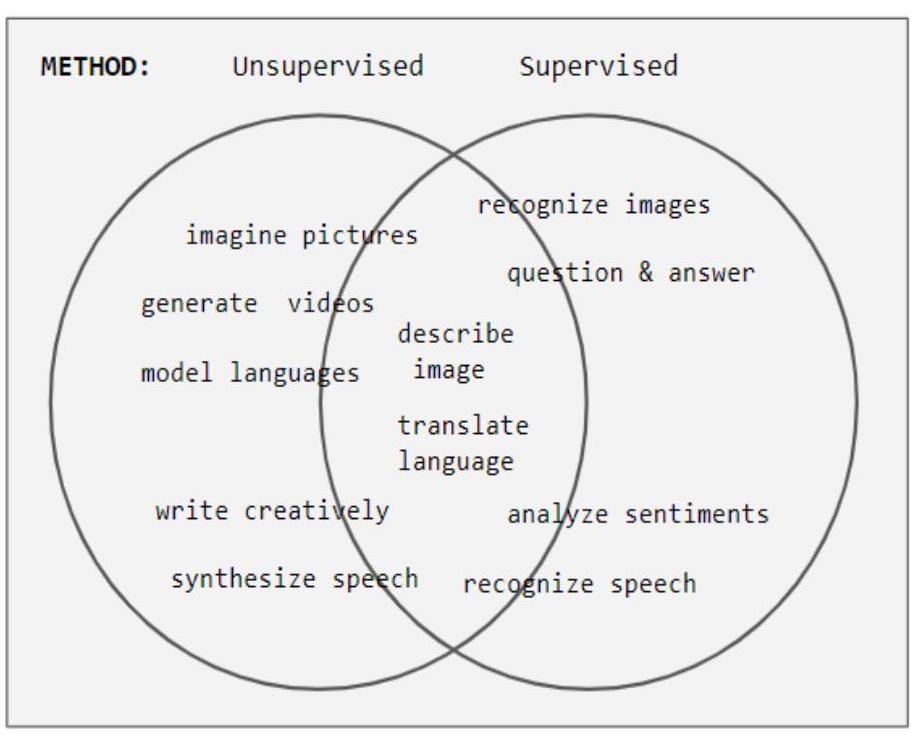}
  \caption{Various changes in supervised and unsupervised models}
\end{figure}
\newline
\newline
In dynamic environments, in which we observed objects moving in unpredictable ways, for example, wandering, sliding, and rotating, the SSL framework was more suitable than the supervised or RL-based methods. In the case of supervised learning models, the problem lies in the fact that they use static and predefined or labelled data, the results of which are not applicable when dealing with dynamic objects on which the model has not been trained. The RL-based method, being more flexible, took a large number of episodes to perform satisfactorily during dynamic environments. But it was able to go up to $65 \%$ even after training for a longer time, after which its performance didn't improve [5].
\newline
\newline
Additional observations in real life also supported the SSL approach's effectiveness. Even in more controlled settings where objects were purposefully placed in new positions or otherwise shifted to replicate real-world interaction, the SSL framework persisted in performing better on average than the baselines [5]. For example, objects left on a tray with smooth linear motion or in a complicated clattering context were better recognized and grasped with SSL. The model was shown to take less time to adapt and have higher levels of grasp success across a wide range of object velocities as well as its movement patterns.

\section*{VII. Conclusion}
In conclusion, a new kind of self-supervised learning approach to meet the challenges of robotic grasping in unstructured environments is proposed. The presented approach allows the robots to learn various aspects of grasping from the vision together with feeling the objects through the proprioceptive feedback without the need to train with the help of labels. Stressful test results show improvements in grasp success rates and learning curves compared to conventional supervised and reinforcement learning. Upcoming work will consider the generalization of the proposed approach towards complex problems as well as into areas of multi-object manipulation and cluttered scenes for real-world applicability.

\section*{References}
\begin{enumerate}
  \item G. Peng, Z. Ren, H. Wang, and X. Li, "A SelfSupervised Learning-Based 6-DOF Grasp Planning Method for Manipulator," IEEE Transactions on Automation Science and Engineering, 2021.
  \item S. Levine, P. Pastor, A. Krizhevsky, J. Ibarz, and D. Quillen, "Learning Hand-Eye Coordination for Robotic Grasping with Deep Learning and Large-Scale Data Collection," The International Journal of Robotics Research, vol. 37, no. 4-5, pp. 421-436, 2017.
  \item Y. Wang, K. Mokhtar, C. Heemskerk, and H. Kasaei, "Self-Supervised Learning for Joint Pushing and Grasping Policies in Highly Cluttered Environments," Frontiers in Robotics and AI, 2022.
\item Wang, Y., Mokhtar, K., Heemskerk, C., \& Kasaei, H. (2024, May). Self-supervised learning for joint pushing and grasping policies in highly cluttered environments. In 2024 IEEE International Conference on Robotics and Automation (ICRA) (pp. $13840-13847)$.
\item J. Mahler et al., "Dex-Net 2.0: Deep Learning to Plan Robust Grasps with Synthetic Point Clouds and Analytic Grasp Metrics," Robotics: Science and Systems, 2017.

\end{enumerate}
\end{document}